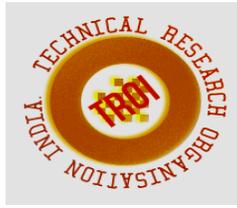

# A COMPREHENSIVE REVIEW ON VARIOUS STATE OF ART TECHNIQUES FOR EYE BLINK DETECTION

Sannidhan M S[1], Sunil Kumar Aithal S[2], Abhir Bhandary[3]

NMAM Institute of Technology

**Abstract**

**Computer Vision is considered to be one of the most important areas in research and has focused on developing many applications that has proved to be useful for both research and societal benefits. Today we have been witnessing many of the road mishaps happening just because of the lack of concentration while driving. As a part of avoiding this kind of disaster happening in day to day life, there are many technologies focusing on keeping track of the vehicle driver's concentration. One such technology uses the method of eye blink detection to find out the concentration level of the driver. With the advent of many high end camera devices with cost effectiveness factor, today it has become more efficient and cheaper to use eye blink detection for keeping track of the concentration level of the driver. Hence this paper presents an exhaustive review on the implementations of various eye blink detection algorithms. The detection system has also extended its application in various other fields like drowsiness detection, fatigue detection, expression detection etc.**

**Index Terms: Computer Vision, Eye blink Detection, Image Processing, Human Computer Interaction, Societal Research.**

## I. INTRODUCTION

Ever increasing road traffic has led to the number of road accidents that is increasing day by day. The cause for many of the accidents is majorly due to two reasons: 1. Due to the negligence and irresponsibility of the drivers. 2. Due to the lack of concentration in driving [1-3]. A driver lacking in the concentration on road is either due to the fatigue or due to the strain in eye. Many research articles have even proved that the over strain in eyesight in some conditions even causes the invisibility for a very short period of time. Apart from causing the invisibility very often the eye strain leads to drowsiness that forces the driver to make a mistake and cause the road mishap by loosing complete control over the vehicle. Research studies have found out that when a person gets into drowsier state, the eye blink rate reduces than the normal value when the person is active [4]. Apart from the usage of this eye blink detection for calculating fatigue level or drowsiness level, currently it has good number of usage in the area of implementing computer vision oriented applications that are controlled just by the sequence of eye blink detection. When it comes to this area of research one good application is meant for the paralyzed people that can control various electronic and electrical devices just through the detection of eye blink sequence [5]. With the advancement in Human Computer Interaction and Computer vision Systems, eye blink detection gained next level of popularity in measuring the emotional expressions of human beings through the dynamics of eye blink detection and sequencing [6-12]. Hence this review paper focuses on the discussion different techniques that detect the eye blink of a person. The aim of this review paper is to discover a efficient and accurate technique for the eye blink detection considering the affordability issues.

Eye blink in simple terms is basically a quick action that promotes the closing and opening of the eyelids. It even has a wide range of applications in human computer interaction. Monitoring the eye blinking and also detecting eye state has proved to be one of the most difficult task in the area of computer vision research since eye occupies a very small region in the entire area of the face [13-15]. Hence in this regard there is also a requirement of using modern technologies that aid in the implementation of eye blink detection easier than the actual traditional approach. Different papers have suggested the use of different modern technologies. Some papers have even suggested the use of wearable entities like goggles, helmets etc that has technology embedded within it

Another major area of research that is currently using eye blink detection to a greater usage is for the finding out the drowsiness factor in individual. Many of the studies have proved the fact that eye blink detection is also on the lethal factors for detecting the drowsiness factor in an individual [16]. Apart from this as a trendy research area today eye blink detection and its sequence is also used as a method of communication between people with disabilities. In the researches reviewed it is being found out that most of the researcher articles have used two separate steps one for eye tracking and another one for eye blink detection [17-18].

## II. AN EXHAUSTIVE REVIEW ON VARIOUS EYE BLINK DETECTION TECHNIQUES

In their research article [1], Choi, I., Han, S., & Kim, D has presented a one-of-a-kind eye tracking and eye detection methodology. The technique mainly treats image as a two dimensional signals that makes use of already defined signal processing methods. In this technique an input is an image captured from a camera or a device like video camera and output might be an image or some important features that is associated with the corresponding input. The entire concept works according to the following four stages: 1. In Pre-processing stage, an input video frame is extracted from the video. The extracted frame is converted into gray color image. 2. In Face Detection stage, The Viola Jones face detector is used as first stage to detect the face along with AdaBoost algorithm. 3. In segmentation stage, an image is basically converted to binary format to apply various morphological features of dilation, area, height and width. In the fourth stage, eye blink detection is carried out using the method of Euclidian distance.

Naveen Kumar H N and Dr. Jagadeesha S in their research article [2] proposed a method for measuring physiological drowsiness by the usage of eye blink detection. In their paper they used the methodology of HOG (Histogram of Oriented Gradient) to extract the yawning state and eye blink detection from the face. As a means of classification of features, they employed the technique of linear SVM (Support Vector Machine) classifier in two different stages. One for eye blink detection and another for yawning state detection. As a method of detecting the level of physiological drowsiness, they used the methods of EEG (Electro Encephalon Graphic) activity, and eye blink duration (eye closure). In their research as part of the implementation model, as part of the face detection approach, Viola Jones method is implemented. One of the main advantages noted for using the aforesaid technique is that it executes faster. The method has considered the pre-processing step as optional but has also proved that the pre-processors improve the detection rate. HOG feature extractor is then employed to capture the detection of eye blink and yawning state. Major limitation identified in the research work is that driver drowsiness is altogether a different drowsiness when compared to a normal drowsiness. Another challenge identified in the proposed research is that eye blinking is also effected by road lighting and oncoming lights. The system also doesn't cover the consideration of wearing of glasses and other wearable's on to the eye.

Jayamala K. Patil and Lego G. Mathew [3] proposed a method which uses IR sensor for eye blink movement detection. The IR rays are transmitted into the driver eyes and the IR detector receives the reflected rays. By continuously monitoring the driver's eye blink rate with the help of IR emitter & detector the microcontroller is fed with this information as input and triggers the alarm system whenever necessary. Microcontroller is connected to LCD which displays the status or alert information. Driver has to use this sensor system as a wearable and non distractible goggle.

Tariq Jamil, Iftaquaruddin Mohammed, and Medhat H. Awadalla [4] implemented an innovative eye blink detector system for automobile accident prevention. For real-time eye-blink detection and monitoring they used

OpenCV tool consisting of functions which focussed on real-time image processing and analysis. Centroid analysis algorithm was used to track the eye blink rate by computing the eye-position within high or low occlusion conditions.. During the car movement the USB enabled camera pre-installed on the car's dashboard continuously monitors the driver's eyes movement. Microcontroller is used to receive the tracked real-time information from the USB camera. Based on this information microcontroller initiates the comparison estimations between the successive readings of the eye blinking information. If any one or both the eyes are found to be closed for a certain time interval, the system alerts the driver by alarm system otherwise system will do nothing. Repetitively, if the buzzer keeps on beeping then the car braking system will get activated. Finally, the concerned authorities are informed about this via alert message.

Atish Udayashankar, Amit R. Kowshik and S. Chandramouli [5] in their research article suggested the techniques for paralyzed eye blink detection. It works according to the following four stages: 1. In Face Detection stage, Haar face detector is used to track and outline the face region. 2. In Pre-initialization stage, after noise removal successive frame differences are estimated along with the count of an eye pair connecting components. A recursive labelling procedure estimates the connecting components and it must be two for an eye pair. 3. In template creation and comparison stage, an open eye related larger connected points are selected for efficiency purpose in order to create a standard template. Using this template eye is properly located based on current eye image matched with this template using square of the difference approach. 4. In Blink tracking stage, eye motion is analyzed and tracked with the help of connected components.

Krystyna Malik and Bogdan Smolka [6] proposed a method for eye blink detection based on local binary patterns. LBP operator produces labelled array information of eye image pattern. First, a histogram samples are generated and compared to detect eye was opened or closed. The LBP histograms represent the uniform distribution of the captured eye region image features. If the difference between histograms tends to be very large then the eye is assumed to be closed in this approach. Also, for proper distance estimation between histograms Bhattacharyya distance and the Kullback-Leibler divergence algorithms are incorporated. Next, for enhancing eye image quality it is de-noised using Savitzky-Golay filter (SGF) which is based on local least-squares polynomial approximation technique is employed. Finally, histograms peak is detected using the morphological top-hat transformation (higher peaks) technique and the strong peaks obtained in the signal are the observation for the identification of eye blinks.

Emiliano Miluzzo, Tianyu Wang and Andrew T. Campbell [7] proposed EyePhone, a novel "hands-free" interfacing system based on Nokia N810/N900 mobile applications/functions using only the user's eyes movement and actions. An example of an EyePhone application is EyeMenu. EyePhone tracks the user's eye movement across the phone's display using the camera mounted on the front of the phone and various machine learning algorithms based on Human-Computer Interaction are used to perform following steps: i) track the eye and infer its position on the Nokia N810/N900 mobile phone display using front camera. Ii) Detect eye blinks that emulate mouse clicks to activate the target application under view. Iii) Eye contour pair is estimated, correlation based template matching technique is used for accurate eye tracking and a thresholding technique for the normalized correlation coefficient returned by the template matching function tracks blink detection.

Hoang Le, Thanh Dang and Feng Liu [8] proposed a method which incorporates an eigen-eye technique for detecting or monitoring the eye-close in the captured individual video frames. Their method learns eye blink patterns and detects eye blinks using a Gradient Boosting algorithm. Next, a non-maximum suppression algorithm is used for consecutive video sequences to remove repeated detection of the same eye-blink action. This approach used a prototyped smart glasses equipped with a low-power camera and an embedded processor. Results with more than 96% accuracy on video frames of a small size of $16 \times 12$ at 96 fps were observed. It finds applications in healthcare, driving safety, and human-computer interaction. Here, instead of purely camera based smart 9horoug with wide range of illumination conditions working capabilities is used.

Liangjun Zhang, Kaiyue Lu, Chengyi Pan and Siyu Xia [9] proposed an idea for communication between user and e-map application through eye movement detection. Here Average filtering and histogram equalization algorithms are used for smoothing of noises and to achieve a greater contrast in the saved image sample. An efficient method called vertical noise reduction for eliminating the interferences of eyelids is used. The improved projection method by them proved to be stable and accurate by the experiment. Thus, accurate eye detection and location tracking, the e-map control application introduced presents a diversity of functions. Movement and zooming of the map and browsing modes are supported by this approach with great accuracy

Amardeep Singh and Amardeep Singh Virk [10] proposed a HUMAN COMPUTER INTERFACE (HCI) systems which was a non recursive system for detecting eye blink of the person during driving. For Real-time eye detection MATLAB's Image processing Vision.CascadeObjectDetector built-in function is used. Also Viola-Jones algorithm can be used for face objects such as human faces, noses, eyes, mouth detection. If the eye remains closed for more than the assumed fixed duration then an alert system is triggered. Initial captured image through camera is gray scale converted. Next, using im2BW() function binary image consisting of 0 for black and 1 for white pixels are generated. Finally, If it found that eye is opened then it will return to for capturing the real time video again.

Narender Kumar and Dr. N.C. Barwar [11] developed an application in C++ using OpenCV tool in Windows environment. The system processes 25 – 30 frames per second for driver's drowsiness detection and tracks yawning in the real time. 1) Face and eye is tracked using Viola Jones method of OpenCV by training set of image samples. 2) An absolute thresholding is done to track eye state. An intensity map is plotted depicting distribution of eyeball pixels on the Y-axis. By monitoring the cliffs of the plotted map height of the eyeball is estimated and the eye state is assessed. 3) Contour finding algorithm is used to detect mouth open or not to predict yawning. 4) if the driver blinks the eyes again and again in a short period of time then the alarm system will get triggered.

The performance of their system was experimented under different lighting conditions without eyeglasses and with eye glasses. More accurate results were 10horough for the driver without eye glasses using this system. The positive alert without eye glasses were recorded for 18-25 age group drivers and the negative or no alert were obtained for the drivers with eye glasses for 50-60 age groups.

Marc Lalonde, David Byrns, Langis Gagnon, Normand Teasdale andDenis Laurendeau [12] work is based on the facial feature of choice for the computation of the cognitive load to detect eye blinks.1) A profile analysis based eye detection is employed for row-wise greyscale features averaging and experiments show x-coordinates are much less stable from frame to frame. 2) Scale-invariant feature transform (SIFT) the GPU-based of OpenVIDIA is used for feature point extraction. 3) Eye Blink is tracked using eye movement detection based on threshold frame differentiation algorithm along with blob filtering.

Chinnawat Devahasdin Na Ayudhya and Thitiwan Srinark [13] suggested Camshift algorithms are efficient for detecting and tracking human face. Adaptive Haar Cascade Classifier was implemented for eye tracking. Eyelid's State Detecting algorithm was proposed for computing accurate threshold value for eye blink detection. Finally, they used the estimated Eyelid's State Detecting values to infer the state of the eyelid and if high Eyelid's State Detecting value is found then the eyes will be in closed state or else the eyes will be in opened state.

Md. Talal Bin Noman and Md. Atiqur Rahman Ahad [14] proposed a real-time system which uses a simple Android mobile phone for tracking human eye blinks. This approach consists of four main stages: (1) Using front camera of android mobile device the real-time image is captured and Haar classifier is used for face detection, (2) Based on the region of interest (ROI) in the frames the proper eye area is extracted, (3) Haar cascade eye tree eyeglasses machine learning boosting classifier algorithm and template-matching approach tracks and detects eye-center; (4) Using the normalized summation of square of difference approach and previous stage online template thorough eye-blink is detected. Finally, an alarm system is triggered based on the detected eye blinks patterns.

ALEKSANDRA KRÓLAK [15] suggested eye-blink detection approach consisting of six stages. It is a system for mental fatigue monitoring purpose and also it is vision-based supporting human-computer interface feature for disabled people who are capable of blinking voluntarily. The six stages are: 1) frames in the captured image sequences are obtained 2) A well known Haar like classifiers are used for face detection 3) Based on some standard geometrical dependencies observed in a human face and incorporating traditional rules of proportion localization of eyes are achieved 4) Normalized template matching technique is employed for only eye region tracking and extraction 5) Repeatedly eye region can be extracted based on above stages whenever a face is tracked 6) Eye-blink is detected and thoroughly analyzed using skin colour segmentation technique and an active standard contour model in YCC colour space.

Junwen Wu and Mohan M. Trivedi [16] proposed particle filters based technique for eye tracking and eye-blink detection. Two filters one for closed eye state and another for open eye state are assumed. For estimating the eyes movement a second-order auto regression model is used. For accuracy a classification-based particle filtering framework for simultaneous eye blink tracking and recognition is incorporated. TensorPCA which is an expansion of the PCA algorithm is used for subspace analysis and feature extraction. For posterior estimation a standard logistic regression model is also used in this approach.

## TABLE I. SUMMARY OF DIFFERENT EYE BLINK DETECTION TECHNIQUES

| Article | Approach | Benefits | Limitations |
|---|---|---|---|
| [1], [10], [11] | Viola Jones Approach | Automatic learning for specified eye appearance | Doesn't work for varying illumination conditions |
| [2] | HOG with linear SVM | High accuracy for the measure of drowsiness | Does not adapt for huge pose variations |
| [3] | Infrared Sensor and detector | Adaptability for the pose variations, greater working distance | Fails to calibrate common eye blink measurement units |
| [4] | Centroid analysis algorithm | Eye blinking detection during night time | Expensive hardware system used |
| [5], [15] | Haar face detector with frame difference system | Better efficiency without face tracking | Lower efficiency for limited lighting conditions |
| [6] | LBP along with Bhattacharyya distance and the Kullback-Leibler divergence algorithms | Higher efficiency rate for varying frame numbers and resolution | Works best for very less number of video samples |
| [7] | Thresholding technique for the normalized correlation coefficient | Capable of tracking eye movement with a low resolution front camera of the mobile phone. | Has resulted in good number of false eye blink contours |
| [8] | Eigen-eye technique for detecting or monitoring the | Eye blink detection is implemented using a very limited hardware | Decrease in performance with the increase in feature values |
|  | eye-close | resource of a smart glass. |  |
| [9] | e-map application based | Efficient eye tracking with geo-location tracking | UI is not user friendly and the work lacks accuracy under some constraints |
| [12] | SIFT and Threshold frame differentiation method | A higher detection rate with lowest execution time | Does not cover the exceptional cases of eye blink detection |
| [13], [14] | Camshift algorithm along with adaptive haar cascade classifier | A dynamic change in blink rate for a specified amount of time is detected and considered | Lower performance and requires a system with higher level of configuration |
| [16] | Tensor PCA | Covers both internal and external environmental factors irrespective of variation in the light | Does not have a technique to improve the images with noise factors |

## III. CONCLUSION

This review paper demonstrates the various methodologies, advantages and disadvantages of eye blink detection under computer vision and image processing areas. Adaptive Haar cascade classifier and TensorPCA based methodologies proved the most widely used and efficient approaches for eye blink tracking and detection. There are good amount of contribution of Haar Face detector and Viola Jones algorithm for various image processing domains that include face detection, proper eye region tracking, facial image processing, simultaneous eye blinks detection, drowsiness and fatigue monitoring and Human Computer Interaction etc.

## IV. REFERENCES

[1] Choi, I., Han, S., & Kim, D. (2011). Eye Detection and Eye Blink Detection Using AdaBoost Learning and Grouping. *2011 Proceedings of 20th International Conference on Computer Communications and Networks (ICCCN).* doi:10.1109/icccn.2011.6005896.

[2] Naveen Kumar H N., & Dr. Jagadeesha S.Physiological Measure of Drowsiness Using Image Processing Technique. IJRASET Vol 4 Issue -VII, July 2016, pp 417-423.

[3] Jayamala K. Patil and Lego G. Mathew, Eye Blinking Monitoring System for Vehicle Accident Prevention.International Journal of Electronic and Electrical Engineering. ISSN 0974-2174 Volume 3, Number 3(2010), pp. 133–138.

[4] T. Jamil, I. Mohammed and M. H. Awadalla, "Design and implementation of an eye blinking detector system for automobile accident prevention," *SoutheastCon 2016*, Norfolk, VA, 2016,pp.1-3.doi: 10.1109/SECON.2016.7506734.

[5] A. Udayashankar, A. R. Kowshik, S. Chandramouli and H. S. Prashanth, "Assistance for the Paralyzed Using Eye Blink Detection," *2012 Fourth International Conference on Digital Home*, Guangzhou, 2012, pp. 104-108.

[6] K. Malik and B. Smolka, "Eye blink detection using Local Binary Patterns," 2014 International Conference on Multimedia Computing and Systems (ICMCS), Marrakech, 2014, pp. 385-390. doi: 10.1109/ICMCS.2014.6911268

[7] Miluzzo, E., Wang, T., & Campbell, A. T. (2010). EyePhone. *Proceedings of the Second ACM SIGCOMM Workshop on Networking, Systems, and Applications on Mobile Handhelds - MobiHeld 10.* doi:10.1145/1851322.1851328

[8] H. Le, T. Dang and F. Liu, "Eye Blink Detection for Smart Glasses," *2013 IEEE International Symposium on Multimedia*,

Anaheim, CA, 2013, pp.305-308. doi: 10.1109/ISM.2013.59

[9] L. Zhang, K. Lu, C. Pan and S. Xia, "Eye Detection for Electronic Map Control Application," *2014 Sixth International Conference on Intelligent Human-Machine Systems and Cybernetics*, Hangzhou, 2014, pp. 241-244.

[10] Amardeep Singh and Amardeep Singh Virk, Real Time Drowsy Driver Identification Using Eye Blink Detection. IJARCSSE, Vol 5, Issue 9, September 2015, pp 335-340.

[11] Narender Kumar and Dr. N.C. Barwar, Detection of Eye Blinking and Yawning for Monitoring Driver's Drowsiness in Real Time, IJAIEM, Vol 3, Issue 11, November 2014, pp. 291-298

[12] M. Lalonde, D. Byrns, L. Gagnon, N. Teasdale and D. Laurendeau, "Real-time eye blink detection with GPU-based SIFT tracking," *Computer and Robot Vision, 2007. CRV '07. Fourth Canadian Conference on*, Montreal, Que., 2007, pp. 481-487. doi: 10.1109/CRV.2007.54

[13] Chinnawat Devahasdin Na Ayudhya, Thitiwan Srinark, "A Method for Real-Time Eye Blink Detection and Its Application", IEEE Computer Society Conf. on Computer Vision and Pattern Recognition (CVPR).

[14] Talal Bin Noman, Md & Ahad, Md. Atiqur Rahman. (2018). Mobile-Based Eye-Blink Detection Performance Analysis on Android Platform. Frontiers in ICT. 5. 4. 10.3389/fict.2018.00004.

[15] A. Krolak and P. Strumillo, "Vision-based eye blink monitoring system for human-computer interfacing," *2008 Conference on Human System Interactions*, Krakow, 2008, pp. 994-998. doi: 10.1109/HSI.2008.4581580

[16] Wu, J. & Trivedi, M.M. "Simultaneous Eye Tracking and Blink Detection with Interactive Particle Filters", EURASIP J. Adv. Signal Process.(2007) 2008: 823695. https://doi.org/10.1155/2008/823695

[17] Ioana Bacivarov, Mircea Ionita, Peter Corcoran, "Statistical models of appearance for eye tracking and eye blink destection and measurement "IEEE transactions on consumer electronics, Vol.54 , No.3, pp. 1312-1320 August 2008.

[18] N. Gregory, C. Thursfield, and D.O. Gorodnichy, "Eye blinks for control. Case study," in Proc. Of 7th European Conference for the Advancement of Assistive Technology, Dublin, Ireland, September (ISBN 1 58603 373 5), 2003.

[19] P. Smith, M. Shah, and N. D. V. Lobo, "Monitoring head/eye motion for driver alertness with one camera," in Proceedings of the International Conference on Pattern Recognition, vol. 15, pp. 636–642, Cambridge, UK, September 2000.

[20] K. Grauman, M. Betke, J. Lombardi, J. Gips, and G. Bradski, "Communication via eye blinks and eyebrow raises: video-based human-computer interfaces,"Universal Access in the Information Society, vol. 2, no. 4, pp. 359–373, 2003.

[21] Sunil Kumar Aithal S and Reeja S R. Article: An Automated System for Detecting Congestion in Huge Gatherings. IJCA Proceedings on International Conference on Information and Communication Technologies ICICT(5):29-32, October 2014.

[22] A. Panning, A. Al-Hamadi and B. Michaelis, "A color based approach for eye blink detection in image sequences," 2011 IEEE International Conference on Signal and Image Processing Applications (ICSIPA), Kuala Lumpur, 2011, pp. 40-45.

[23] N. Kojima, K. Kozuka, T. Nakano and S. Yamamoto, "Detection of consciousness degradation and concentration of a driver for friendly information service", pp. 31-36

[24] S. Agrawal, I. Constandache, S. Gaonkar, and R.R. Choudhury. PhonePoint Pen: Using Mobile Phones to Write in Air. In MobiHeld '09,pages1–6.ACM,2009.

[25] J. Liu, L. Zhong, J. Wickramasuriya, and V. Vasudevan. uWave: Accelerometer-Based Personalized Gesture Recognition and its Applications. In Pervasive and Mobile Computing,vol. 5, issue 6, pp. 657-675, December 2009.

[26] E. Miluzzo, C.T. Cornelius, A. Ramaswamy, T. Choudhury, Z. Liu, A.T. Campbell. Darwin Phones: The Evolution of Sensing and Inference on Mobile Phones. In Eighth International ACM Conference on Mobile Systems, Applications, and Services (MobiSys '10), San Francisco, CA, USA, June 15-18, 2010

[27] Xiaobo Guo. Eye-Controlled Mouse System Research Based On EyeTracking Technology[D]. Tianjin University, 2009.

[28] Viola P, Jones M. Rapid object detection using a boosted cascade of simple features[C].Computer Vision and Pattern Recognition, 2001. CVPR 2001. Proceedings of the 2001 IEEE Computer Society Conference on. IEEE, 2001, 1: I-511-I-518

[29] Topal, C., Gerek, Ö. N. and Doğan, A., "A head-mounted sensor-based eye tracking device: eye touch system," Proc. of the 2008 Symp. on Eye tracking research & applications, 2008, pp. 87-90.

[30] Huang, G. B., Ramesh, M., Berg, T. and Learned-Miller, E., "Labeled Faces in the Wild: A Database for Studying Face Recognition in Unconstrained Environments," Technical Report 07-49, University of Massachusetts, Amherst, October, 2007.

[31] Heishman, R. and Duric, Z., "Using Image Flow to Detect Eye Blinks in Color Videos," Proc. of the IEEE Workshop on Applications of Computer Vision (WACV), Austin, Texas, USA, February, 2007.

[32] S. Han, S. Yang, J. Kim, and M. Gerla, "Eyeguardian: a framework of eye tracking and blink detection for mobile device users." in HotMobile. 13th Workshop on Mobile Computing Systems and Applications 2012, G. Borriello and R. K. Balan, Eds. ACM, 2012, p. 6.

[33] S. Madhav, P. Gaddamwar, and G. Chipte, "Eyegaze communication system," International Journal of Computer & Communication Technology (IJCCT), vol. 3, pp. 33–37, 2012

[34] K. Radlak and B. Smolka, "Blink detection based on the weighted gradient descriptor," in Proceedings of the 8th Int. Conf. on Computer Recognition Systems CORES 2013, ser. Advances in Intelligent Systems and Computing. Springer, 2013, vol. 226, pp. 691–700.

[35] S. Krishnan and C. Seelamantula, "On the selection of optimum Savitzky-Golay filters," Signal Processing, IEEE Transactions on, vol. 61, no. 2, pp. 380–391, 2013.

[36] P. S. Parmar and N. Chitaliya, "Detect eye blink using motion analysis method," International Journal of Emerging Technologies in Computational and Applied Sciences (IJETCAS), vol. 4, pp. 180–185, 2013.

[37] Y. Sun, S. Zafeiriou, and M. Pantic, "A hybrid system for online blink detection," in Forty-Sixth Annual Hawaii International Conference on System Sciences, Maui, Hawaii, 2013.

[38] A. A. Lenskiy and J.-S. Lee, "Driver's eye blinking detection using novel color and texture segmentation algorithms," International Journal of Control, Automation and Systems, vol. 10, no. 2, pp. 317–327, 2012.

[39] A. O. Selvı, A. Ferıkoğlu, D. Güzel and E. Karagöz, "Design and implementation of EEG signal based brain computer interface for eye blink detection," 2017 International Conference on Computer Science and Engineering (UBMK), Antalya, 2017, pp. 544-548.

[40] D. Batulin, A. Popov, A. Bobrov and A. Tretiakova, "Eye blink detection for the implantable system for functional restoration of orbicularis oculi muscle," 2017 Signal Processing Symposium (SPSympo), Jachranka, 2017, pp. 1-4.

[41] Z. A. Haq and Z. Hasan, "Eye-blink rate detection for fatigue determination," 2016 1st India International Conference on Information Processing (IICIP), Delhi, 2016, pp. 1-5.

[42] J. S. Jennifer and T. S. Sharmila, "Edge based eye-blink detection for computer vision syndrome," 2017 International Conference on Computer, Communication and Signal Processing (ICCCSP), Chennai, 2017, pp. 1-5.

[43] Kokila, R., M. S. Sannidhan, and Abhir Bhandary. "A study and analysis of various techniques to match sketches to Mugshot photos." Inventive Communication and Computational Technologies (ICICCT), 2017 International Conference on. IEEE, 2017.

[44] R. Kokila, M. S. Sannidhan and A. Bhandary, "A novel approach for matching composite sketches to mugshot photos using the fusion of SIFT and SURF feature descriptor," 2017 International Conference on Advances in Computing,

Communications and Informatics (ICACCI), Udupi, 2017, pp. 1458-1464.

[45] Sannidhan, M. S., and G. Ananth Prabhu. "A Comprehensive Review on Various State-Of-The-Art Techniques for Composite Sketch Matching."

[46] Fernandes, Steven Lawrence, and Josemin G. Bala. "Recognizing faces when images are corrupted by varying degree of noises and blurring effects." Emerging ICT for Bridging the Future-Proceedings of the 49th Annual Convention of the Computer Society of India (CSI) Volume 1. Springer, Cham, 2015.

[47] Fernandes, Steven Lawrence, and G. Josemin Bala. "Time Taken for Face Recognition under varying Pose, Illumination and Facial Expressions based on Sparse Representation." International Journal of Computer Applications 51.12 (2012).

[48] Fernandes, Steven, and Josemin Bala. "A comparative study on various state of the art face recognition techniques under varying facial expressions." Int. Arab J. Inf. Technol. 14.2 (2017): 254-259.

[49] Fernandes, Steven Lawrence, and G. Josemin Bala. "Recognizing Faces in Corrupted Images." Proceedings of the International Conference on Soft Computing Systems. Springer, New Delhi, 2016.

[50] L. Pauly and D. Sankar, "A novel method for eye tracking and blink detection in video frames," 2015 IEEE International Conference on Computer Graphics, Vision and Information Security (CGVIS), Bhubaneswar, 2015, pp. 252-257.